\documentclass[10pt,journal,compsoc]{IEEEtran}
\usepackage{amsmath, amsfonts, amsthm, amssymb}
\usepackage{mathpazo}
\usepackage{enumitem}
\usepackage{graphicx}
\usepackage{multirow}
\usepackage{geometry}
\usepackage{subfig}
\usepackage{hyperref}
\usepackage{algorithm}
\usepackage{algorithmic}
\usepackage{footmisc}
\usepackage{rotating}
\theoremstyle{definition}

\newcommand{\pr}{\text{Pr}}
\ifCLASSOPTIONcompsoc
  \usepackage[nocompress]{cite}
\else
  \usepackage{cite}
\fi

\ifCLASSINFOpdf
\else
\fi

\begin{document}
%
\title{Predictive Modeling with Delayed Information: a Case Study in E-commerce Transaction Fraud Control}
%
%
%
%

\author{Junxuan~Li,
        Yung-wen~Liu,
        Yuting~Jia,
        Yifei~Ren,
        Jay~Nanduri
\IEEEcompsocitemizethanks{\IEEEcompsocthanksitem Junxuan Li is with Microsoft Corporation and H. Milton Stewart School of Industrial \& Systems Engineering, Georgia Institutes of Technology. 755 Ferst Dr NW, Atlanta, GA 30318. Email: junxuxan.li@gatech.edu.\protect \\ \IEEEcompsocthanksitem Yifei Ren is with Department of Mathematics and Computer Science, Emory University. 400 Dowman Dr, Atlanta, GA 30307. Email: yifei.ren2@emory.edu\protect\\ \IEEEcompsocthanksitem Yung-wen Liu, Yuting Jia and Jay Nanduri are with Microsoft Corporation. 4500 154th Pl NE, Redmond, WA 98052. Email: (yungliu, yutjia, jayna)@microsoft.com.}
\thanks{Manuscript received 0; revised 0.\protect\\(Corresponding author: Jay Nanduri.)}}

\IEEEtitleabstractindextext{%
\begin{abstract}
In Business Intelligence, accurate predictive modeling is the key for providing adaptive decisions. We studied predictive modeling problems in this research which was motivated by real-world cases that Microsoft data scientists encountered while dealing with e-commerce transaction fraud control decisions using transaction streaming data in an uncertain probabilistic decision environment. The values of most online transactions related features can return instantly, while the true fraud labels only return after a stochastic delay. Using partially mature data directly for predictive modeling in an uncertain probabilistic decision environment would lead to significant inaccuracy on risk decision-making. To improve accurate estimation of the probabilistic prediction environment, which leads to more accurate predictive modeling, two frameworks, Current Environment Inference (CEI) and Future Environment Inference (FEI), are proposed. These frameworks generated decision environment related features using long-term fully mature and short-term partially mature data, and the values of those features were estimated using varies of learning methods, including linear regression, random forest, gradient boosted tree, artificial neural network, and recurrent neural network. Performance tests were conducted using some e-commerce transaction data from Microsoft. Testing results suggested that proposed frameworks significantly improved the accuracy of decision environment estimation.
\end{abstract}

\begin{IEEEkeywords}
Business intelligence, Data mining, Machine learning, Deep learning, Knowledge acquisition.
\end{IEEEkeywords}}

\maketitle

\IEEEdisplaynontitleabstractindextext

%
\IEEEpeerreviewmaketitle

\IEEEraisesectionheading{\section{Introduction}\label{sec:introduction}}

%
%
%
%
\IEEEPARstart{B}{usiness} Intelligence studies have attracted tremendous interests from both academic research and industrial practice. Every decision is made within a decision environment which is defined as the assemblage of all exogenous responses and activities that affect the loss/reward gained by endogenous decision actions. 

Thanks to developments in big data and machine learning, data-driven decision support systems are able to record streaming data on the cloud, get awareness of decision environment and act accordingly to achieve optimal system performances. Decision environment characterizations are usually not available to decision makers and need to be inferred from data collected through different types of equipment, recorded in a variety of database streams and combined on the cloud platform. Obtaining ideal characterization of the decision environment is costly, sometimes impossible, which lead to a constrained estimation of the decision environment. Machine learning, as one of the major technology set in data mining, provides a plenty of methods to recognize the pattern of decision environment and predicts current decision environment using point or statistical estimations. Adaptive machine learning further enables the decision maker to obtain awareness of environment on-the-fly when he/she chooses proper decision actions.

\begin{figure*}[!t]
\centering
\includegraphics[width=4in]{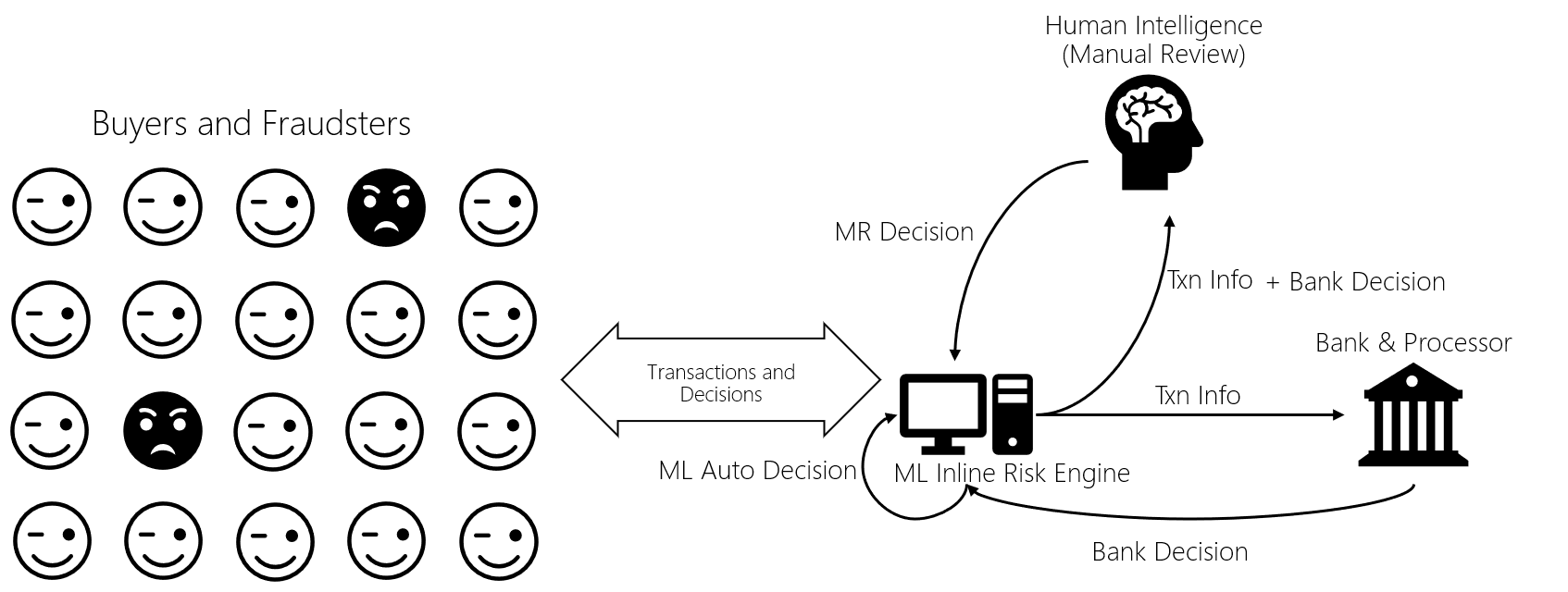}
\caption{Dynamic decision environment demonstration\cite{DynamicFraudControlSystem2018DSS}}
\label{fig:DSS}
\end{figure*}

E-commerce, as one of the major components of modern business, is under threats from unforeseeable online fraudsters. In the paper \cite{DynamicFraudControlSystem2018DSS}, some optimization models built for Microsoft's dynamic e-commerce transaction fraud control system were discussed.  This paper, \cite{DynamicFraudControlSystem2018DSS}, pointed out that fraud control decisions of e-commerce merchants should not be made independently, but interactively with other external associated decision parties, such as banks and manual review agent teams. Transaction information could be only partially shared among different decision parties due to information confidentiality and privacy, e.g. Microsoft could not release customers purchase history to banks, and a bank should not share cardholders' purchase history with other merchandises to Microsoft. Figure \ref{fig:DSS} depicts the dynamic and interactive decision environment of fraud control decision support system (DSS). After a purchase transaction occurs, the risk scoring engine first evaluates the risk level of this transaction using an estimated risk score (transactions with the same risk score are considered as one category of control objectives), and the DSS provides a control action that either approve, review or reject this purchase request. In other words, the task of DSS is to decide how to assign control actions to control objectives in different score categories. Approved transactions are sent to the payment issuing bank for authorization check, among which only bank authorized transactions can be granted final purchase approvals. Reviewed transactions are first sent to the payment issuing bank for authorization check and those bank authored transactions are then sent to manual review agents for further risk screening. Only both bank authorized and manually approved transactions are marked as final approved transactions. If a purchase request receives a rejection from any of the three decision parties, this transaction will be declined and marked as rejected transactions. 

Microsoft had observed some rapid fluctuations in the decision behavior patterns of banks and manual review teams. For example, for transactions belong to the same risk category (having the same risk score), bank approval rate could sometime variate hugely time to time. The decision behavior patterns of banks and manual review teams are dynamic and are highly correlated to the decision quality of fraud control DSS. The behaviors patterns observed are list below:
\begin{itemize}
\item If the recently reported fraud incidence number increases, banks and manual review agents become more conservative in approving purchases transactions, which leads to higher rejection rates;
\item On the other hand, if recently reported fraud incidence number decreases, banks and manual review agents then become less conservative, which results in lower rejection rates but likely to bypass more fraudulent transactions in consequence.
\item In addition, when fewer fraudulent transactions were submitted to manual review teams, since fraud patterns are less massive and obvious, MR teams have more difficult time to detect frauds
\end{itemize}

From the list above, it is not hard to see the importance of studying the interactive behavior patterns  among decision parties with respect to fraud control actions of the e-commerce merchants to achieve the profit optimality.

One of the most commonly used transaction fraud labels in e-commerce is ``chargeback'' which is the return of funds to the credit card holder, initiated by the payment instrument issuing bank to settle a debt.  With this type of fraud label, it is fairly common that the true feedback of decision is inaccessible right away after fraud control decision is made. The delay of fraud labels for purchase transaction is due to the fact that it usually takes some amount of time for the credit card legitimate holders to realize that their cards are misappropriated and file dispute to their bank. In this case, business intelligent models are not able to capture the most recent fraud patterns and to provide the most accurate decision for fraud control. This delay lead time is usually referred as data maturity lead time in big data era.  Inaccurate risk decision made by business intelligent models would cause bad looping effects which fluctuate accuracy of decisions made by all decision-making parties in the decision environment and lead to the less profitable outcome for e-commerce merchants. To be able to estimate decision environment as well as its interaction effect, and make it as an input for the model is critical for making the risk decision system to reach the most profitable decisions.  For fraud control DSS, if we ignore recent data and only use mature data to estimate behavior patterns of other decision parties, our control policy will be already outdated due to the lag of pattern recognition. If we ignore data maturity lead time and use partially mature labels to estimate behavior patterns of other decision parties, we may be mislead by recent decision quality, and in return overestimate/underestimate approval rates of banks or manual review agents. Predictive modeling of the decision environment thus becomes challenging due to delayed information.

The demand for learning decision environment, as well as the challenge of data maturity lead time, provide strong motivation for the authors to design proper inference methods to predict the rapidly fluctuating decision environment for Microsoft e-commerce fraud control DSS. This paper proposes two frameworks, Current Environment Inference (CEI) and Future Environment Inference (FEI) frameworks, that resolve maturity lead time issues in decision environment prediction. Both frameworks first generate decision environment related features using long-term mature data and short-term partially mature data, and estimate decision environment using variety of learning methods, including linear regression (LR), random forest (RF), gradient boosted tree (GB), artificial neural network (ANN) and recurrent neural network (RNN). CEI module is designed to use partially mature data (delayed information) to predict the decision environment of the coming decision epoch. And, FEI module is designed to further estimate decision environment of a future decision epoch to help evaluate the effect of current control decisions in the future decision epoch. Although these frameworks are designed for fraud control DSS, it can be easily customized for the use of other industrial applications that also face challenges of data mining with delayed information.

This paper is organized as the following. Section \ref{sec:review} includes a number of literature that addressed research related to this research topic. Section \ref{sec:partial} first illustrates the structure of partially mature streaming data, then defines decision environment in fraud control. Section \ref{sec:framework} illustrates two frameworks that predict decision environment patterns for fraud control DSS. Readers who are interested in the use of these two frameworks may refer to \cite{DynamicFraudControlSystem2018DSS} for more system control operation details. The Performance of these two frameworks are tested through implementing them on a portfolio of transaction data from Microsoft e-commerce database, and all testing results are discussed in Section \ref{sec:test}. Section \ref{sec:conclusion} concludes the paper and briefly introduces how to extend the use of proposed frameworks to other industries that also face similar challenges of having only partially mature data.

\section{Related Literature}\label{sec:review}
Predictive modeling of decision environment started from the early 1980's.  The paper, \cite{DynamicDecisionMaking1982Payne}, described and emphasized the importance that decision environment should be estimated dynamically and decisions should then be made accordingly. The value of decision environment estimation for a dynamic medical decision-making problem was studied with a simulated medical system in \cite{MedicalDEI1987}. Results in \cite{MedicalDEI1987} suggested that decision behaviors were influenced by the features of decision environment. Following these research, the paper, \cite{RL1998}, proposed probabilistic representations for decision environment prediction and  probabilistic predictions are then updated and reinforced using sequential reward/loss returns from the environment. While on the other research branch, the paper \cite{IBLearning2003} proposed instance-based learning to predict decision environment with the help of similarity-based exemplary database constructed using historical data. Based on the knowledge of the authors, there was currently no existing literature on predictive modeling for e-commerce transaction fraud control. The lack of literature in this field is because of two reasons: (1) It is not easy for academic researchers to have the access to e-commerce transaction data, as these data are strictly confidential; (2) Conventional fraud control models do not consider the interaction effect of decisions made by other decision parties and assume decision environment patterns are fixed.

Suppose a dynamic decision environment can be described using a number of attributes whose values vary as time changes, we can record series of attribute values with respect to time. If at a given time point, the decision environment can be characterized by $n$ attributes, then the values of these attributes can be represented by an $n$ dimensional trajectory. In this way, dynamic decision environment prediction problem can be modeled as a trajectory prediction problem that has rich literature. Trajectory prediction study started from the 1920's \cite{timeseries1927} using classic time series analysis methods. The type of linear prediction methods for time series data, as a special kind of trajectory, are summarized in \cite{timeseriesbook1994}. Recent trajectory prediction researches motivated by different application considered higher trajectory dimension with more features as well as the nonlinear relations between these features, and they adopted machine learning (artificial neural network \cite{elecloadforecast1991}\cite{elecloadnn1992}\cite{annpredictionsurvey}\cite{tsnncomp1996} , random forest \cite{FRlanguage2007}\cite{watersupplyforecast2010}\cite{electricityforecast2015}\cite{trajprediction2014health}\cite{TS-FR-comparison2017}, gradient boosted tree \cite{gb2004}) and deep lLearning (\cite{tsnncomp1996}\cite{rnn2010language}\cite{rnnspeech2011}\cite{rnnlocation2016} \cite{rnnlocation2017}) methodologies. The use of artificial neural network in trajectory prediction started from early 90's motivated by electricity load prediction (e.g. \cite{elecloadforecast1991} \cite{elecloadnn1992}). These paper used neural network to model nonlinear relations between electricity load in past periods (trajectory history) and other features, such as temperature and location. Both papers,\cite{elecloadforecast1991} and \cite{elecloadnn1992}, claimed high prediction accuracy with neural network models. The paper, \cite{annpredictionsurvey} provided a comprehensive review of other applications using neural network in predicting trajectory type data. The paper, \cite{tsnncomp1996}, compared performance of classic time series models with neural network model on real-world price trajectory data, and claimed a significant improvement with neural network model over ARIMA models by reducing the mean square error by 27 - 56 percent. The papers,\cite{FRlanguage2007} and \cite{gb2004}, considered a speech trajectory and adopted random forest and gradient boosted Tree methods, respectively, to predict next trajectory value which was associated with a wording database to predict the next input word. The paper, \cite{watersupplyforecast2010}, adopted random forest method in demand prediction for a water supply system. They constructed long-term demand trend feature, short-term demand calibration feature, and other incidence features to increase prediction accuracy of coming water demand based on historical demand trajectory of different locations. Similar idea was widely used in engineering and medical field (see \cite{electricityforecast2015} for electricity demand trajectory prediction, and \cite{trajprediction2014health} for decease outbreak incidence trajectory prediction). A recent research in \cite{TS-FR-comparison2017} conducted extensive amount of numerical performance comparisons among predictive modeling using classic time series forecasting (ARMA and ARIMA) with different parameters and random forest on 16,000 simulated and 135 real temperature trajectories.  The paper, \cite{TS-FR-comparison2017}, observed that random forest method outperformed the traditional time series methods in most of their tests.
\cite{tsnncomp1996} is one of the pioneer articles that introduced recurrent neural network to trajectory prediction. Recurrent neural network methods exploited temporal dependencies in a time sequence and uses internal states to model interactions between different time steps of the trajectory data. \cite{rnn2010language} and \cite{rnnspeech2011} claimed that recurrent neural network model provided faster and more accurate speech trajectory predictions than traditional artificial neural network. The papers, \cite{rnnlocation2016} and \cite{rnnlocation2017}, adopted recurrent neural network to next location prediction using trajectory history data and user-dependent attributes.

Our research departs from current trajectory prediction research, as what it was mentioned earlier the e-commerce transaction streaming data have delayed labels so that the trajectory cannot represent the exact decision environment history for the periods of recent immature streaming data. Handling delayed information and data mining with delayed labels is considered one of the most important open challenges in big data era \cite{openchalleges2014KDD}. The paper, \cite{openchalleges2014KDD}, identified gaps between current research and meaningful applications, and highlighted the importance and the challenge of predictive modeling using streaming data with some delay in data maturity. Despite the fact that prediction modeling with delayed data immaturity is an important problem, there are only a few literature discussing how to solve it. The papers, \cite{KNNDelayLabel2008}, \cite{TKDEcluster2011} and \cite{SemiSupervisedClustering2012}, used semi-supervised $K$ nearest neighborhood method to resolve clustering problems when cluster labels are partially observed. However, we have not yet found any literature that address regression problem and continuous valued time series prediction problem with delayed information.

\section{Partially mature Data and Fraud Control Decision Environment}\label{sec:partial}
In this section, we first give an overview of the structure of streaming data set collected for fraud control decision engine in Section \ref{sec:data-structure}. Next in Section \ref{sec:decision-environment}, we define the mathematical form of a decision environment.
\subsection{Streaming Data Structure}\label{sec:data-structure}
\begin{figure}[h]
\centering
\includegraphics[width=2.8in]{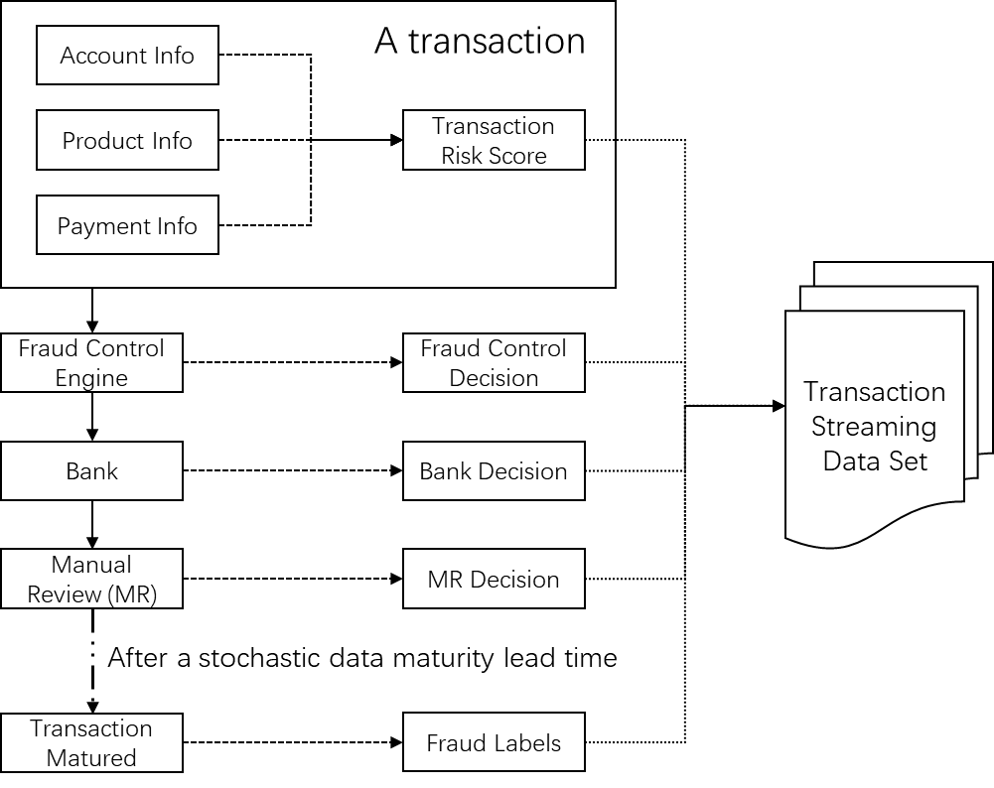}
\caption{Transaction streaming data collection demonstration}\label{fig:streaming-data}
\end{figure}

Figure \ref{fig:streaming-data} demonstrates the structure of streaming data set.

A transaction carries user's account information (e.g. Microsoft account information), product information (e.g. a Surface book with a set of specifications, the total price and cost), and payment information (e.g.  type of payment instrument, location of payment, etc.). The time of occurrence of a purchase transaction is immediately recorded as "ReceivingTime". A risk scoring engine then scores this transaction and record it in the risk control database as "RiskScore".  A fraud control engine then makes control decision which is then recorded as "InlineDecision". The payment issuing bank and the manual review (MR) team make decisions and their decisions are recorded in the database as "BankDecision" and "MRDecision". The fraud label of this transaction is set to "False" by default in "FraudFlag" in the database, and after a stochastic lead time in data maturity, we receive the final fraud label and update transaction's FraudFlag to "True" if a chargeback returns.  The maturity time of a transaction is also recorded simultaneously in the column "MaturityTime".

We discretize the time span into equal length periods, e.g. treat one week as a period. Then transactions occurred in the same period, e.g. same week, are treated as occuring at the same time stamp. Let $L$ be the maximum data maturity lead time (measured in number of periods), then at the beginning of a given period $t$,  the available streaming data can be separated into two segments:
\begin{enumerate}
\item Long-term mature data: streaming data with time stamp no later than $t-L$;
\item Short-term partially mature data: streaming data with time stamp from $t-L+1$ to $t-1$.
\end{enumerate}

\subsection{Decision Environment of Dynamic Fraud Control}\label{sec:decision-environment}
The decision environment of the inline fraud control is characterized by probabilistic measures of Banks' and MRs' decision behavior patterns. Decision environment characteristics are introduced by \cite{DynamicFraudControlSystem2018DSS} in the form of conditional probabilities. Let $s$ denote a risk score of a transaction, where $s$ has a finite integral support $S=\{s_1,s_2,...,s_{|S|}\}$, and the decision environment of an optimal transaction fraud control is characterized by the following five probabilistic functions:
\begin{itemize}
\item Probability that a transaction with score $s$ is authorized by the payment issuing bank and turns out to be non-fraudulent: 
\begin{align*}
&g_1(s)=\pr(\text{Bank Auth.} \cap \text{Non-fraud} \mid s )\\
&=\frac{\text{\# of bank auth. and non-fraud score $s$ trans.}}{\text{\# of finally approved score $s$ trans.}};
\end{align*}
\item Probability that a transaction with score $s$ is authorized by the bank and turns out to be fraudulent:
\begin{align*}
&g_2(s)=\pr(\text{Bank Auth.} \& \text{Fraud} \mid s)\\
&=\frac{\text{\# of bank auth. and fraud score $s$ trans.}}{\text{\# of finally approved score $s$ trans.}};
\end{align*}
\item Probability that a transaction with score $s$ is authorized by the bank and approved by the manual review team, and finally turns out to be non-fraudulent:
\begin{align*}
&g_3(s)=\pr(\text{Bank Auth.} \cap \text{MR App.} \cap \text{Non-fraud} \mid s)\\
&=\frac{\text{\# of bank auth., MR app, non-fraud score $s$ trans.}}{\text{\# of finally approved score $s$ trans.}};
\end{align*}
\item Probability that a transaction with score $s$  is authorized by the bank and approved by the manual review team, and finally turns out to be fraudulent:
\begin{align*}
&g_4(s)=\pr(\text{Bank Auth.} \cap \text{MR App.} \cap \text{Fraud} \mid s)\\
&=\frac{\text{\# of bank auth. , MR app., fraud score $s$ trans.}}{\text{\# of finally approved score $s$ trans.}};
\end{align*}
\item probability that a transaction with score $s$ is authorized by the bank:
\begin{align*}
&g_5(s)=\pr(\text{Bank Auth.}\mid s)\\
&=\frac{\text{\# of bank auth. score $s$ trans.}}{\text{\# of finally approved score $s$ trans.}}.
\end{align*}
\end{itemize}
These five probabilistic functions are called $g$-functions which are short for "gold functions", since their values describe profit related probabilities associated with different risk operations in transaction fraud control system. $g_5(\cdot)$ can be estimated using the most recent transaction streaming data, as bank decision signals are available instantly (within few seconds). While on the  other hand, predicting $g_1(\cdot)$, $g_2(\cdot)$, $g_3(\cdot)$ and $g_4(\cdot)$ are not trivial, since we do not have the up-to-date fraud labels due to data maturity lead time. We will focus on the problem of how to predict $g_1(\cdot)$ to $g_4(\cdot)$ in this paper. We choose predictive modeling of $g_1(\cdot)$ as an example, while estimating the other three $g$ functions follows exactly the same procedures. Let $g_1^t(\cdot)$ be the $g_1$ function in period $t$, then decision environment inference tasks can be stated as the following:
\begin{enumerate}
\item At the beginning of a given period, e.g. period $t$, what is the current decision environment $g^{t}(s)$ for all $s$?
\item During a given period $t$, will the series of control decisions made so far affect $g$ function in the future? Is there a way to estimate future $g$ function, e.g. $g^{t+l}(s)$ for period $t+l$? 
\end{enumerate}

\section{Predictive Modeling Frameworks}\label{sec:framework}
In this section, we illustrate the details of two predictive modeling frameworks that resolve two major tasks proposed in Section \ref{sec:decision-environment}. We propose a general framework for current period decision environment inference in Section \ref{sec:cei}, named as CEI. And considering the fact that decision actions taken so far in current period, e.g. $t$, will affect future decision environment, e.g. period $t+l$, we introduce the second inference framework called FEI in Section \ref{sec:fei}. We use $g_1(\cdot)$ as an example through out this section, and estimating other $g$ functions follows the exact same procedure.

\subsection{Current Environment Inference (CEI) Framework}\label{sec:cei}
\begin{figure}[h]
\centering
\includegraphics[width=3in]{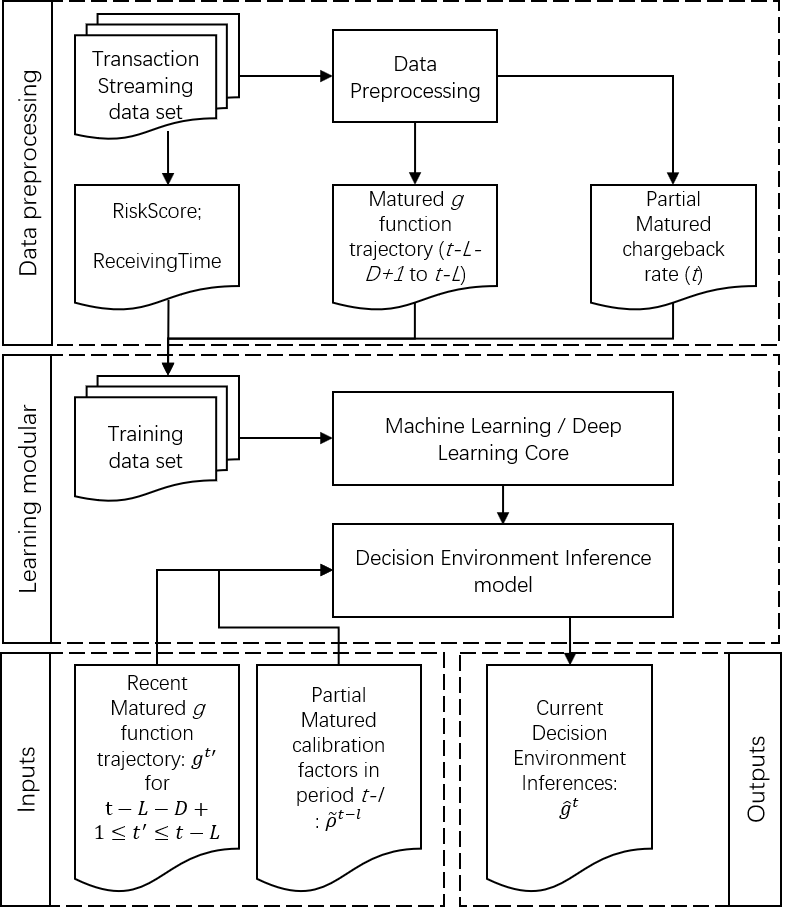}
\caption{CEI framework demonstration}
\label{fig:cei}
\end{figure}

Figure \ref{fig:cei} depicts the system logics for Current Environment Inference (CEI) framework.
CEI framework consists two segments, the data pre-processing and the learning module. As the size of transaction streaming data set expands by each period, CEI model is updated at the beginning of each decision period. Most updated transaction streaming data set is first pre-processed into a training data set, and machine learning/deep learning method can be deployed to build a model for current decision environment inference. In Section \ref{sec:obtain-feature}, we first demonstrate how to obtain useful features and construct training data for CEI. Pre-processed training data will be later fed into learning module in Section \ref{sec:cei-learn} and produce a inference model that maps input features to an prediction of current period $g_1$ function.

\subsubsection{Data Pre-processing}\label{sec:obtain-feature}
Recall that decision environment in period $t$ is characterized by $g_1^t(s)$ for all $s\in S$. Then at any given risk score $s$, the values of $g_1^t(s)$ compose a time series along the time axis. Considering the fact that $g_1^t(s_1)$ might be correlated with $g_1^t(s_2)$ for $s_1\neq s_2$ for any period $t$, we shall include risk score $s$ and period number $t$ as two features of training data in the first place.

We introduce Long-Term-Short-Term (LTST) idea since we collect features at different time points,  that would help improve estimation accuracy. Recall that the lead time in data maturity is at most $L$ periods, so that at the beginning of a given decision period $t$, we have access to exact decision environment for any periods earlier than $t-L$. In this way, we include the most recent mature $D$ decision environment information, i.e. $g_1^{t-L-D+1}(\cdot)$, $ g_1^{t-L-D+2}(\cdot)$ ,..., $g_1^{t-L}(\cdot)$, as the features of training data. On the other hand, given the fact that the bank and the MR team always adjust their decision behavior patterns as described in Section \ref{sec:introduction}, we calculate biased chargeback rates in recent periods using partial mature streaming data. We do not have the access to the chargeback rates $\rho^{t'}$ in period $t'\in\{t-L+1,...,t-1\}$, but the biased chargeback rate $\tilde{\rho}^{t'}$ can be estimated as
$$\frac{\left(\begin{array}{c}
	\text{\# of chargeback transactions}\\ \text{in week $t'$ occurred before week $t$}
	\end{array}\right)}{\text{(\# of finally approved transactions in week $t'$)}}.$$
Using correlation tests, we are able to find out if any of these biased chargeback rates have connections with $g_1^t(\cdot)$. We include the most related $\tilde{\rho}^{t'}$ as the features of training data. For example, if $\tilde{\rho}^{t-l}$ has significant correlation with $g_1^t$, we include this biased chargeback rate in period $t-l$ into the feature set of training data. Responses of the training data are the exact $g_1^t(s)$ values.

The idea of constructing this training data is adopted from trajectory prediction research. For each risk score $s$, $g_1^t(s)$ at a series of periods, i.e. a series of$t$, can be considered as a trajectory. Furthermore, we consider the fact that for any different $s$, the trajectory of $g_1$ function might be correlated and should not be estimated separately. The intuition of including the most recent mature $D$ decision environment information, i.e. $g_1^{t-L-D+1}(\cdot)$, $ g_1^{t-L-D+2}(\cdot)$ ,..., $g_1^{t-L}(\cdot)$, as features of training data is to record the most recent available exact trajectory to estimate the level and trend of long-term $g_1$ function. While on the other hand, including a recent biased chargeback rate $\tilde{\rho}^{t-l}$ in the feature set of training data provides a short-term calibration factor to amend the long-term $g_1$ function estimation so that our $g_1$ function estimations are more representable to reflect the most recent decision environment. We summarize data structure of CEI training in Table \ref{tbl:cei-data}, where "x" indicates the training feature data and "o" indicates the training response data.

\begin{table}[h]
\centering\caption{CEI Training Data Summary}
\scalebox{0.85}{
\begin{tabular}{|c|c|c|c|c|c|c|c|}
\hline
Week & $t-L-D+1$ & $\cdots$ & $t-L$ & $\cdots$ & $t-l$ & $\cdots$ & $t$ \\ \hline
$g$ function & x & x & x &  &  &  & o  \\ \hline
\begin{tabular}[c]{@{}c@{}}Partial\\ chargeback\\ rate\end{tabular} &  &  &  &  & x &  &  \\ \hline
\end{tabular}}
\label{tbl:cei-data}
\end{table}

\subsubsection{CEI Learning module}\label{sec:cei-learn}
We consider the learning module as a regression problem that maps input features to an estimation of $g_1$ function of current period. A number of alternative methods can be considered as the core learning models. We provide readers a variety of options in machine learning and deep learning, including linear regression (LR), artificial neutral network (ANN), random forest (FR), gradient boosted tree (GB), and recurrent neutral network (RNN). With the model trained (and model parameters are tuned using cross validation) at the beginning of the decision period $t$, most recent $D$ mature $g_1$ functions, i.e. $g_1^{t'}(s)$ for $t-L-D+1\le t'\le t-L$ at all $s\in S$, and calibration factor, i.e. $\tilde{\rho}^{t-l}$, composes the input, and CEI model outputs estimation of $g_1$ functions in period $t$ for all $s\in S$, which is denoted by $\hat{g}_1^t(s)$,
\begin{align*}
\hat{g}_1^t(s)=\Phi(g_1^{t'}(s),\tilde{\rho}^{t-l}:t-L-D+1\le t'\le t-L).
\end{align*}

The scope of this paper is to illustrate a general framework for decision environment inference with partial mature data. In this way, we actually leave certain degree of freedom for readers to choose a more suitable learning method (out of LR, ANN, RF, GB and RNN, or using other regression based learning methods with linear or nonlinear structure). We compare the results of performance tests for CEI with LR, ANN, RF, GB and RNN as the learning cores in Section \ref{sec:test}.

\subsection{Future Environment Inference (FEI) Framework}\label{sec:fei}
\begin{figure*}[h]
\centering
\includegraphics[width=5in]{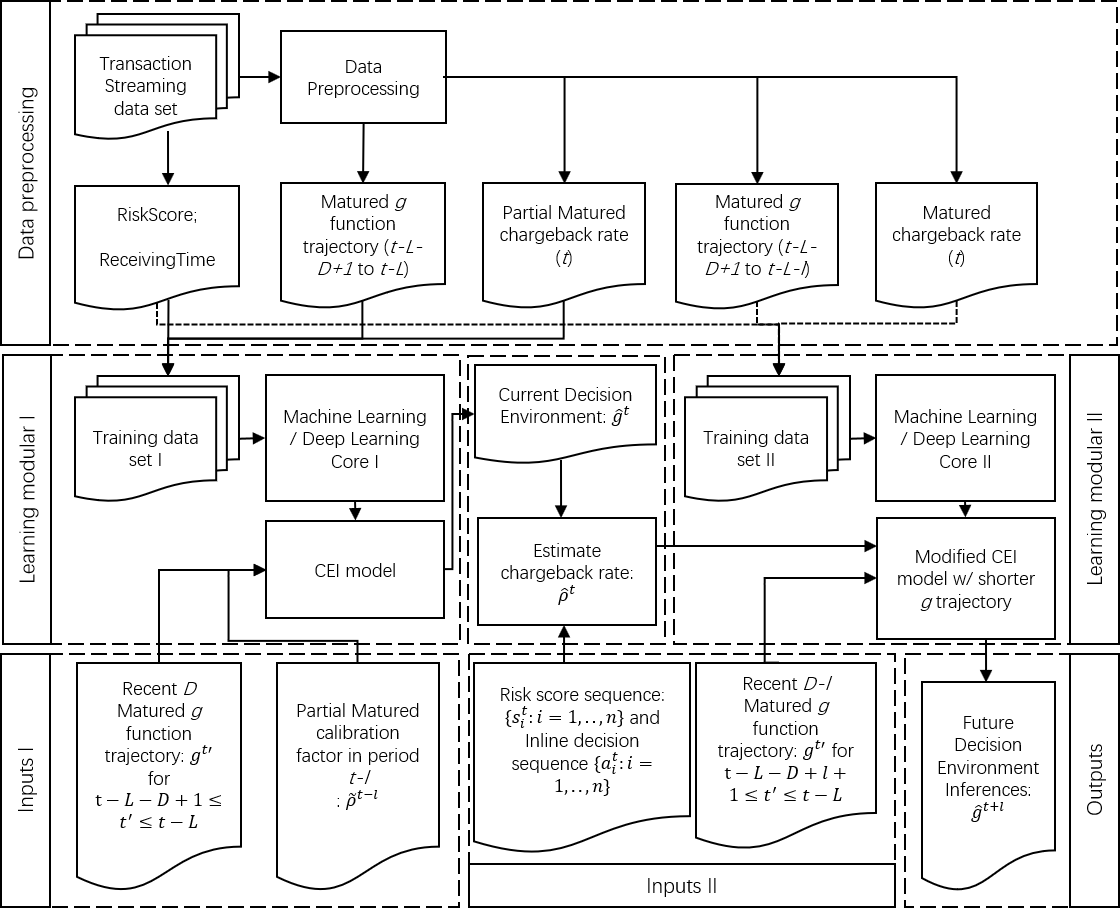}
\caption{FEI framework demonstration}
\label{fig:fei}
\end{figure*}

Figure \ref{fig:fei} depicts system logic for Future Environment Inference (FEI) framework.
FEI framework consists a data pre-processing module and two learning modules. Same as CEI, FEI model is also updated once per period, after including new streaming data and updating fraud labels. However, FEI has the following two main differences from CEI.
\begin{enumerate}
\item Data processing module of FEI transforms transaction streaming data into two training data sets: Training data set I includes $g_1$ function trajectories of length $D$, and contributes in training learning module I; while training data set II includes shorter $g_1$ function trajectories of length $D-l$, and will be used by learning module II;
\item Two training data sets are fed into two learning modules: learning module I is identical with CEI learning module； and learning module II is a modified version of CEI learning module with a shorter $g_1$ function trajectory for training and prediction.
\end{enumerate}
Details of data pre-processing and learning modules are introduced in Section \ref{sec:obtain-feature-2} and \ref{sec:fei-learning} respectively.

\subsubsection{Data Pre-processing}\label{sec:obtain-feature-2}
In this section, we will illustrate what are considered as useful features to predict future decision environment, and how to pre-process streaming data into training data sets.

Similar as CEI data pre-processing, we introduce LTST idea to include the potential features observed at different time points. CEI suggests that the $D$ most recent mature $g_1$ function trajectory being put into training feature set.  For example, for the period $t+l$ long-term trend of $g_1$ function should be estimated from $g^{t+l-L-D+1}_1(\cdot)$ to $g^{t+l-L}_1(\cdot)$. However, at the beginning of the decision period at $t_0$, we only have access of $g^{t_0+l-L-D+1}_1(\cdot)$ to $g^{t_0-L}_1(\cdot)$. In this way, we use $g^{t_0+l-L-D+1}_1(\cdot)$ to $g^{t_0-L}_1(\cdot)$ to estimate long-term trend of $g^{t+l}_1(\cdot)$, so that this shorter trajectory of length $D-l$ should be included into feature set. CEI suggests that chargeback rate in period $t$, i.e. $\rho^t$, is correlated with $g^{t+l}_1(\cdot)$ which should be included into feature set. This feature set contributes to pre-process training data II in Figure \ref{fig:fei}.

Training data II is not enough to predict decision environment at period $t+l$, because for the time being that we predict $g^{t+l}_1$ in the decision period, i.e. during period $t$, we do not know the chargeback rate $\rho^t$. However, if we have access to $g^t_2(s)$ and $g^t_4(s)$, for transaction sequence occurred in current period so far (actions from the beginning of the decision period to the time we make $g^{t+l}_1$ estimations), let $\{a^t_i:i=1,2,...,m, a^t_i\in\{App.,Rev.,Rej.\}\}$ be the action sequence, and $\{s^t_i:i=1,2,...,m\}$ be the risk score sequence, we are able to obtain an estimation of $\rho^t$,
\begin{align*}
\hat{\rho}^t=\frac{1}{\sum_{i=1}^{m}\delta_{(a^t_i\neq Rej.)}}\left(\sum_{i=1}^m g^t_2(s^t_i)\cdot\delta_{(a^t_i=App.)}\right.\\\hfill\left.+\sum_{i=1}^m g^t_4(s^t_i)\cdot\delta_{(a^t_i=Rev.)} \right).
\end{align*}
where $\delta_{(H)}$ is indicator function of event $H$, i.e. $\delta_{(H)}=1$ if $H$ is true and $\delta_{(H)}=0$ otherwise. As Section \ref{sec:cei} provide methods to estimate $g^t_2(s)$ and $g^t_4(s)$, we need to utilize CEI model to first obtain estimations of $g^t_2(s)$ and $g^t_4(s)$, i.e. $\hat g^t_2(s)$ and $\hat g^t_4(s)$, and use estimated $g^t_2(s)$ and $g^t_4(s)$ to calculate $\hat{\rho}^t$. In this way, we construct training data set I of FEI using exactly the same way that we construct training data set for CEI.
\begin{table*}[ht]
\centering\caption{FEI Training Data Summary}
\scalebox{0.85}{
\begin{tabular}{|c|c|c|c|c|c|c|c|c|c|c|c|}
\hline
Week & \begin{tabular}[c]{@{}c@{}}$t-L-$\\ $D+1$\end{tabular} & $\cdots$ & \begin{tabular}[c]{@{}c@{}}$t-L-D$\\ $+l+1$\end{tabular} & $\cdots$ & $t-L$ & $\cdots$ & $t-l$ & $\cdots$ & $t$ & $\cdots$ & $t+l$ \\ \hline
g function & x $\Delta$ & x $\Delta$ & x $\Delta$ & x & x &  &  &  & o &  & $\square$ \\ \hline
\begin{tabular}[c]{@{}c@{}}Partial\\ chargeback\\ rate\end{tabular} &  &  &  &  &  &  & x &  &  &  &  \\ \hline
\begin{tabular}[c]{@{}c@{}}Chargeback\\ rate\end{tabular} &  &  &  &  &  &  &  &  &  $\Delta$ &  &  \\ \hline
\end{tabular}}
\label{tbl:fei-data}
\end{table*}
We summarize data structure of FEI training in Table \ref{tbl:fei-data}. Training data set I consists feature data with "x" tags and response data with "o" tag. Training data set II consists feature data with "$\Delta$" tags and response data with "$\square$" tag.

\subsubsection{FEI Learning modules}\label{sec:fei-learning}
FEI framework includes two learning modules. Learning module I produces a model that maps the most recent $D$ mature $g$ function trajectory, i.e. $g^{t'}(s)$ for $t-L-D+1\le t'\le t-L$ at all $s\in S$, and calibration factor, i.e. $\tilde{\rho}^{t-l}$, to $g$ function estimations, $\hat{g}^t_j(\cdot)$ for $j\in\{1,2,4\}$. With $\hat{g}^t_2(\cdot)$ and $\hat{g}^t_4(\cdot)$, we can estimate chargeback rate in the current decision period $t$ , i.e. $\rho^t$. Learning module II then is used to produce a model that maps a shorter recent mature $g$ function trajectory, i.e. $g^{t'}(s)$ for $t-L-D+l+1\le t'\le t-L$ at all $s\in S$, and calibration factor, i.e. $\hat{\rho}^{t}$, to $g_1$ function estimation in period $t+l$, $\hat{g}^{t+l}_1(\cdot)$.
\begin{align*}
\hat{g}_1^{t+l}(s)=\Psi(g_1^{t'}(s),\hat{\rho}^{t}:t-L-D+l+1\le t'\le t-L).
\end{align*}

Similar with CEI framework, FEI framework can use a variety of learning cores for leaning module I and II. We suggest several options for both learning cores, including linear regression (LR), artificial neutral network (ANN), random forest (FR), gradient boosted tree (GB), and recurrent neutral network (RNN). We compare results of performance tests for FEI with LR, ANN, RF, GB and RNN as the learning cores in Section \ref{sec:test}.

\section{A Case Study of Microsoft E-commerce}\label{sec:test}
In this section, we conduct systematic performance tests for CEI and FEI modules using real-world e-commerce transaction data from Microsoft. For the chosen business, we consider one decision period as one week, and previous research in Microsoft has concluded the maximum data maturity lead time to be $L=12$ weeks, meaning that at week $t$, all transactions occurred in or before week $t-12$ has exact fraud labels. Hence all $g$ functions have the exact function value in any period less or equal to $t-12$. We choose most recent one month  $g$ functions to estimate long-term $g$ function trends, i.e. $D=4$ weeks. Next we show how to choose the proper chargecback and partial chargeback rate as short-term calibration factor in a statistical way.

With mature historical data, we can obtain score aggregated value of $g$ function for each week, for instance,
\begin{align*}
&g_1^t=\pr(\text{Trans. is Bank Auth.} \cap \text{Non-fraud in $t$})\\
&=\frac{\text{\# of bank auth. and non-fraud trans. in week $t$}}{\text{\# of finally approved trans. in week $t$}};
\end{align*}
We calculate full chargeback rate and partial chargeback rate $l$ weeks ago, e.g. full chargeback rate $\rho^{t-l}$ is calculated as $$\frac{\text{(\# of total chargeback transactions in week $t-l$)}}{\text{(\# of finally approved transactions in week $t-l$)}},$$ and $l$ week partial chargeback rate $\tilde \rho^{t-l}$ is calculated as $$\frac{\left(\begin{array}{c}
	\text{\# of chargeback transactions}\\ \text{in week $t-l$ occurred before week $t$}
	\end{array}\right)}{\text{(\# of finally approved transactions in week $t-l$)}}.$$
We believe that current $g^t$ is correlated with $\rho^{t-l}$ and $\tilde \rho^{t-l}$. 

We use $g_1$ as a demonstration example. We collect values of $g^t_1$ and $\rho^{t-l}$ from historical data, where $l=1,2,3,4$. Then we conduct non-parametric statistical tests to claim the correlation and the monotonic relation between $g^t_1$ and $\rho^{t-l}$. Testing results are summarized in Table \ref{tbl:test-rho-g}.

\begin{table}[h]
	\centering	\caption{Tests result summary: $g^t_1$ v.s. $\rho^{t-l}$}
	\scalebox{0.9}{\begin{tabular}{|c|c|c|c|c|}
\hline
$l$ & 1 & 2 & 3 & 4 \\ \hline
Kendall correlation & -0.474 & -0.486 & -0.462 & -0.433 \\ \hline
Kendall p-val & 0.022 & 3.071e-4 & 9.017e-5 & 0.003 \\ \hline
Spearman correlation & -0.646 & -0.690 & -0.704 & -0.625 \\ \hline
Spearman p-val & 0.016 & 3.297e-4 & 1.541 e-4 & 0.007 \\ \hline
\end{tabular}}
\label{tbl:test-rho-g}
\end{table}

\begin{figure}[htbp]
\centering
\subfloat[$g^t_1$ v.s. $\rho^{t-2}$]{\includegraphics[width=2.5in]{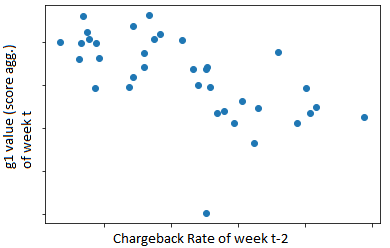}}\\
\subfloat[$g^t_1$ v.s. $\rho^{t-3}$]{\includegraphics[width=2.5in]{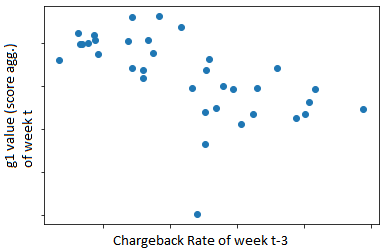}}
\caption{Vitalization of $g^t_1$ v.s. $\rho^{t-l}$, $l=2,3$}\label{fig:relation1}
\end{figure}

The p-values of Kendall tests suggest that we can reject Kendall null hypothesis ($H_0$: $g^t_1$ and $\rho^{t-l}$ are independent), and hence we can claim the significant weekly dependence between $g^t_1$ and $\rho^{t-l}$ for $l=2,3$. Spearman tests also verify the monotonically decreasing trend of $g^t_1$ and $\rho^{t-l}$, especially for $l=2,3$, as the p-values of Spearman tests suggest that Spearman null hypothesis is rejected ($H_0$: $g^t_1$ and $\rho^{t-l}$ do not have monotone increasing/decreasing relation). Monotone decreasing trends of $g^t_1$ v.s. $\rho^{t-2}$ and $g^t_1$ v.s. $\rho^{t-3}$ is visualized in Figure \ref{fig:relation1}.

\begin{figure} [htbp]
\centering
\includegraphics[width=2.8in]{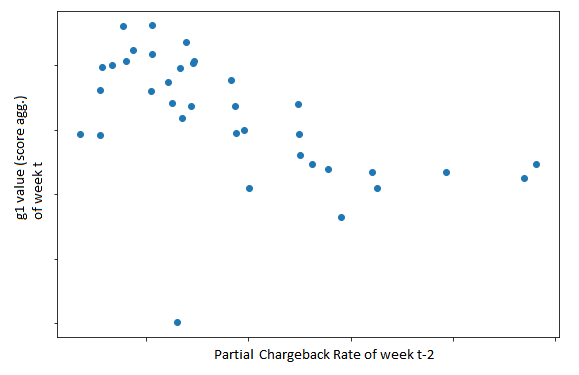}
\caption{Vitalization of $g^t_1$ v.s. $\tilde \rho^{t-2}$}\label{fig:relation2}
\end{figure}

We choose $l=2$ and include $\rho^{t-2}$ as a short-term calibration factor for its smallest Kendall and Spearman p-values. Similarly we test correlation and monotonic relation between $g^t_1$ and $\tilde \rho^{t-2}$, in order to have $\tilde \rho^{t-2}$ included as a short-term calibration factor. Results in Table \ref{tbl:test-prho-g} confirm dependence between $g^t_1$ and $\tilde \rho^{t-2}$, as well as monotone decreasing trend of $g^t_1$ and $\tilde \rho^{t-2}$. This decreasing monotone relation is validated as shown in Figure \ref{fig:relation2}.

\begin{table}[htbp]
	\centering\caption{Tests result summary:  $g^t_1$ v.s. $\tilde \rho^{t-2}$}
	\begin{tabular}{|c|c|c|}
		\hline
		& Correlation & P-value  \\ \hline
		Kendall tau test & -0.467      & 6.21e-4 \\ \hline
		Spearman r test  & -0.639      & 2.72e-4  \\ \hline
	\end{tabular}
	\label{tbl:test-prho-g}
\end{table}

We conducted 10 weeks of field test with Microsoft e-commerce transaction streams. We also conducted performance tests of CEI and FEI framework in parallel with a number of learning cores including linear regression (LR), artificial neural network (ANN), random forest (RF), gradient boosted tree (GB), and recurrent neural network (RNN). We recorded our predictions of $g$ functions for all $s\in S$ for the 10 testing weeks. True $g$ functions were calculated 12 weeks later and we reported mean square errors (MSE) and error standard deviations of different versions of CEI and FEI in Table \ref{tbl:cei-results} and Table \ref{tbl:fei-results}.

\begin{table*}[htbp]
	\centering\caption{CEI framework performance test result summary}
	\scalebox{1.0}{\begin{tabular}{|c|c|c|c|c|c|}
\hline
CEI Learning Core & Performance Measure & $g_1$ & $g_2$ & $g_3$ & $g_4$ \\ \hline
\multirow{2}{*}{LR} & MSE & 0.0086 & 0.0029 & 0.0033 & 0.0014 \\ \cline{2-6} 
 & Error Std. & 0.0189 & 0.0078 & 0.0058 & 0.0018 \\ \hline
\multirow{2}{*}{ANN} & MSE & 0.0062 & 0.0025 & 0.0037 & 0.0016 \\ \cline{2-6} 
 & Error Std. & 0.0105 & 0.0063 & 0.0076 & 0.0020 \\ \hline
\multirow{2}{*}{RF} & MSE & 0.0081 & 0.0027 & 0.0025 & 0.0011 \\ \cline{2-6} 
 & Error Std. & 0.0165 & 0.0015 & 0.0052 & 0.0015 \\ \hline
\multirow{2}{*}{GB} & MSE & 0.0086 & 0.0032 & 0.0028 & 0.0011 \\ \cline{2-6} 
 & Error Std. & 0.0185 & 0.0087 & 0.0059 & 0.0015 \\ \hline
\multirow{2}{*}{RNN} & MSE & 0.0075 & 0.0021 & 0.0029 & 0.0009 \\ \cline{2-6} 
 & Error Std. & 0.0136 & 0.0064 & 0.0052 & 0.0013 \\ \hline
\end{tabular}}
\label{tbl:cei-results}
\end{table*}
\begin{table*}[htbp]
	\centering\caption{FEI framework performance test result summary}
	\scalebox{1}{\begin{tabular}{|c|c|c|c|c|c|}
\hline
FEI Learning Core & Performance Measure & $g_1$ & $g_2$ & $g_3$ & $g_4$ \\ \hline
\multirow{2}{*}{LR} & MSE & 0.0098 & 0.0036 & 0.0033 & 0.0023 \\ \cline{2-6} 
 & Error Std. & 0.0145 & 0.0068 & 0.0046 & 0.0016 \\ \hline
\multirow{2}{*}{ANN} & MSE & 0.0068 & 0.0277 & 0.0098 & 0.0187 \\ \cline{2-6} 
 & Error Std. & 0.0145 & 0.0184 & 0.0156 & 0.0122 \\ \hline
\multirow{2}{*}{RF} & MSE & 0.0097 & 0.0040 & 0.0037 & 0.0026 \\ \cline{2-6} 
 & Error Std. & 0.0155 & 0.0074 & 0.0052 & 0.0021 \\ \hline
\multirow{2}{*}{GB} & MSE & 0.0095 & 0.0039 & 0.0037 & 0.0019 \\ \cline{2-6} 
 & Error Std. & 0.0202 & 0.0082 & 0.0056 & 0.0029 \\ \hline
\multirow{2}{*}{RNN} & MSE & 0.0079 & 0.0030 & 0.0038 & 0.0013 \\ \cline{2-6} 
 & Error Std. & 0.0136 & 0.0058 & 0.0058 & 0.0014 \\ \hline
\end{tabular}}
\label{tbl:fei-results}
\end{table*}

We first discuss details of CEI framework testing results. From Table \ref{tbl:cei-results}, we observed that CEI with ANN had the smallest MSE as well as the smallest error standard deviation for $g_1(\cdot)$ over all scores. However for $g_2(\cdot)$, $g_3(\cdot)$, and $g_4(\cdot)$, RNN learning core outperformed all other methods by providing the smallest MSE. Moreover, by compare error standard deviations, we can claim that CEI-RNN performance was relatively robust for all $g$ functions, as it always yielded the smallest or second smallest error standard deviations. CEI-ANN had the largest error standard deviations for predicting $g_2(\cdot)$, $g_3(\cdot)$, and $g_4(\cdot)$, which indicates its instability in predictive modeling. We can claim CEI-RF the most robust prediction methods by providing the smallest error standard deviations and relatively small MSE's. CEI with LR and GB had mediocre performances in this group of parallel testing study.

FEI testing results in Table \ref{tbl:fei-results} show the fact that double estimation of FEI framework did increase prediction MSE's, as the inputs of learning moduld II are related to outputs of learning module I which could be inaccurate. In this test, FEI with RNN learning core had the best performance by providing the smallest MSE's, as well as the smallest standard deviations for all $g$ functions. CEI-ANN had the worst performance as its MSE's for $g_2(\cdot)$, $g_3(\cdot)$, and $g_4(\cdot)$ were much larger than all other methods. FEI frameworks with LR, RF and GB had similar performance, while FEI-GB yielded slightly smaller MSE's and FEI-LR had relatively smaller error standard deviations.

Lastly, we discuss computational performance of CEI and FEI framework. Algorithm designers should care about both prediction accuracy and computational complexity of each version of CEI and FEI frameworks. A good predictive modeling framework should not only provide accurate decision environment predictions, but also has the acceptable computational time for training the learning module(s) to derive a prediction, i.e. output of $\Phi(\cdot)$ functions and $\Psi(\cdot)$ function.

Table \ref{tbl:comp-results} summarized the average computation time and its standard deviation for each learning module (e.g. CEI learning module, FEI learning module I and FEI learning module II) with each learning core (e.g. LR, ANN, RF, GB, RNN) based on parallel tests on 100 training samples using the same personal computer with Intel Core i7-7700HQ CPU. Each computation time included cross validation time and model training time. LR method had the shortest computational time in average with less than 1 millisecond. ANN had the second shortest computational time with a few milliseconds. RF and GB methods have similar computational performance. Obtaining a prediction using RF and GB as learning core required 80 to 95 milliseconds. Training a RNN based learning module required around 10 seconds which was the cost of accurate prediction. We provide these computational time results for algorithm designers in Microsoft e-commerce risk control group. Algorithm designers may sometime face the trade-off between prediction accuracy and computational complexity.

\begin{table}[htbp]
	\centering\caption{Computational result summary}
	\scalebox{1}{\begin{tabular}{|c|c|c|c|}
\hline
\begin{tabular}[c]{@{}c@{}}Learning\\ Core\end{tabular} & \begin{tabular}[c]{@{}c@{}}Comp.\\ Time\end{tabular} & \begin{tabular}[c]{@{}c@{}}CEI \&\\ FEI \\ module I\end{tabular} & \begin{tabular}[c]{@{}c@{}}FEI\\ module II\end{tabular} \\ \hline
\multirow{2}{*}{LR} & Avg.(s) & 6.1e-4 & 5.0e-4 \\ \cline{2-4} 
 & Std.(s) & 8.8e-4 & 7.0e-5 \\ \hline
\multirow{2}{*}{ANN} & Avg.(s) & 5.5e-3 & 9.7e-3 \\ \cline{2-4} 
 & Std.(s) & 9.4e-4 & 1.2e-3 \\ \hline
\multirow{2}{*}{RF} & Avg.(s) & 7.7e-2 & 7.5e-2 \\ \cline{2-4} 
 & Std.(s) & 4.7e-3 & 7.4e-3 \\ \hline
\multirow{2}{*}{GB} & Avg.(s) & 8.9e-2 & 9.3e-2 \\ \cline{2-4} 
 & Std.(s) & 5.3e-3 & 5.0e-3 \\ \hline
\multirow{2}{*}{RNN} & Avg.(s) & 9.8 & 10.1 \\ \cline{2-4} 
 & Std.(s) & 3.4 & 3.7 \\ \hline
\end{tabular}}
\label{tbl:comp-results}
\end{table}

\section{Conclusion}\label{sec:conclusion}
We conclude our research by highlighting contributions in this paper, and by introducing how the use of proposed frameworks can be extended to other industries with the challenges of having only partially mature data.

In this paper, we discussed about predictive modeling with delayed information based on a real world application in e-commerce transaction fraud dynamic control. Two frameworks , Current Environment Inference (CEI) and Future Environment Inference (FEI) frameworks, are proposed to resolve the issue of long lead time in data maturity in the fraud control decision environment prediction. These two frameworks construct prediction features using long-term-short-term idea, and obtain long-term trend features using mature data and short-term calibration features using partially mature data. A number of learning methods are also proposes as candidates for learning core of both framework, including linear regression, Random Forest, Gradient Boosted Tree, Artificial Neural Network, and Recurrent Neural Network. Performance tests were conducted on a portfolio of e-commerce transaction data from Microsoft to compare different versions of CEI and FEI. Testing results suggest that proposed frameworks have a great prediction accuracy. We had observed the great potential of using Recurrent Neural Network in predictive modeling with delayed information. However, if the predictive modeling requires millisecond level computational time, we would suggest Random Forest or Gradient Boosted Tree as the candidates to be first considered by algorithm designers.

The ideas behind CEI and FEI modules can be easily adopted and extended to other industries with delayed information, such as sales prediction for inventory control, citation prediction for journal ranking, multi-sensor recognition with delayed signals, and etc.. Instead of using up-to-date data to learn the most recent fraud patterns for predictions, we can keep track of long-term trend of the predicting target for the long-term estimations, and find a correlated short-term factor to calibrate predictions based on the long-term estimation. By assuming linear or nonlinear relations between the long-term factors, the short-term factors and the predicting target, data scientists are able to train machine learning and deep learning models using the mature data set. In this way, the use of proposed frameworks can be generalized to a more boarder category of applications in predictive modeling with delayed information.

\ifCLASSOPTIONcompsoc
\section*{Acknowledgments}
\else
\section*{Acknowledgment}
\fi
This research was supported by Microsoft, Seattle, WA. The authors are
thankful to members in Membership Knowledge and Growth group at Microsoft for data technical support
and any anonymous reviewers and referees for their constructive commons.


%



\ifCLASSOPTIONcaptionsoff
  \newpage
\fi



\bibliographystyle{IEEEtran}
\bibliography{IEEEabrv,mybibfile}

\begin{thebibliography}{10}
\providecommand{\url}[1]{#1}
\csname url@samestyle\endcsname
\providecommand{\newblock}{\relax}
\providecommand{\bibinfo}[2]{#2}
\providecommand{\BIBentrySTDinterwordspacing}{\spaceskip=0pt\relax}
\providecommand{\BIBentryALTinterwordstretchfactor}{4}
\providecommand{\BIBentryALTinterwordspacing}{\spaceskip=\fontdimen2\font plus
\BIBentryALTinterwordstretchfactor\fontdimen3\font minus
  \fontdimen4\font\relax}
\providecommand{\BIBforeignlanguage}[2]{{%
\expandafter\ifx\csname l@#1\endcsname\relax
\typeout{** WARNING: IEEEtran.bst: No hyphenation pattern has been}%
\typeout{** loaded for the language `#1'. Using the pattern for}%
\typeout{** the default language instead.}%
\else
\language=\csname l@#1\endcsname
\fi
#2}}
\providecommand{\BIBdecl}{\relax}
\BIBdecl

\bibitem{DynamicFraudControlSystem2018DSS}
\BIBentryALTinterwordspacing
J.~Li, Y.~Liu, Y.~Jia, and J.~Nanduri, ``Discriminative data-driven
  self-adaptive fraud control decision system with incomplete information,''
  \emph{arXiv:1810.01982 [cs.AI]}, 2018. [Online]. Available:
  \url{https://arxiv.org/pdf/1810.01982.pdf}
\BIBentrySTDinterwordspacing

\bibitem{DynamicDecisionMaking1982Payne}
J.~W. Payne, ``Contingent decision behavior,'' \emph{Psychological Bulletin},
  vol.~92, no.~2, pp. 382--402, 1982.

\bibitem{MedicalDEI1987}
D.~N. Kleinmuntz and J.~B. Thomas, ``The value of action and inference in
  dynamic decision making,'' \emph{Organizational Behavior and Human Decision
  Processes}, vol.~39, pp. 341--364, 1987.

\bibitem{RL1998}
R.~S. Sutton and A.~G. Barto, \emph{Reinforcement Learning—An
  Introduction}.\hskip 1em plus 0.5em minus 0.4em\relax Cambridge, MA: MIT
  Press, 1998.

\bibitem{IBLearning2003}
C.~Gonzalez, J.~F. Lerch, and C.~Lebiere, ``Instance-based learning in dynamic
  decision making,'' \emph{Cognitive Science}, vol.~27, p. 591–635, 2003.

\bibitem{timeseries1927}
G.~U. Yule, ``On a method of investigating periodicities in disturbed series
  with special reference to wolfer’s sunspot numbers,'' \emph{Philosophical
  Transactions of the Royal Society London}, no. 226, pp. 267--298, 1927.

\bibitem{timeseriesbook1994}
G.~E.~P. Box, G.~M. Jenkins, and G.~C. Reinsel, \emph{Time Series Analysis:
  Forecasting and Control}.\hskip 1em plus 0.5em minus 0.4em\relax NJ:
  PrenticeHall: Englewood Cliffs, 1994.

\bibitem{elecloadforecast1991}
D.~Park, M.~El-Sharkawi, R.~Marks, L.~Atlas, and M.~Damborg, ``Electric load
  forecasting using an artificial neural network,'' \emph{IEEE Transactions on
  Power Engineering}, vol.~6, no.~2, pp. 442--449, 1991.

\bibitem{elecloadnn1992}
K.~Y. Lee, Y.~T. Cha, and J.~Park, ``Short-term load forecasting using an
  artificial neural network,'' \emph{Transactions on Power Systems}, vol.~7,
  no.~1, pp. 124--132, 1992.

\bibitem{annpredictionsurvey}
G.~Zhang, B.~E. Patuwo, and M.~Y. Hu, ``Forecasting with artificial neural
  networks: The state of the art,'' \emph{International Journal of
  Forecasting}, no.~14, pp. 35--62, 1998.

\bibitem{tsnncomp1996}
N.~Kohzadi, M.~S. Boyd, B.~Kermanshahi, and I.~Kaastra, ``A comparison of
  artificial neural network and time series models for forecasting commodity
  prices,'' \emph{Neurocomputing}, no.~10, pp. 169--181, 1996.

\bibitem{FRlanguage2007}
P.~Xu and F.~Jelinek, ``Random forests and the data sparseness problem in
  language modeling,'' \emph{Computer Speech and Language}, vol.~21, pp.
  105--152, 2007.

\bibitem{watersupplyforecast2010}
M.~Herrera, L.~Torgo, J.~Izquierdo, and R.~Perez-Garcia, ``Predictive models
  for forecasting hourly urban water demand,'' \emph{Journal of Hydrology},
  vol. 141-150, no. 387, 2010.

\bibitem{electricityforecast2015}
G.~Dudek, \emph{Advances in Intelligent Systems and Computing}.\hskip 1em plus
  0.5em minus 0.4em\relax Springer, Cham, 2015, vol. 323, ch. Short-Term Load
  Forecasting Using Random Forests.

\bibitem{trajprediction2014health}
M.~J. Kane, N.~Price, M.~Scotch, and P.~Rabinowitz, ``Comparison of arima and
  random forest time series models for prediction of avian influenza h5n1
  outbreaks,'' \emph{BMC Bioinformatics}, vol.~15, no. 276, 2014.

\bibitem{TS-FR-comparison2017}
H.~Tyralis and G.~Papacharalampous, ``Variable selection in time series
  forecasting using random forests,'' \emph{Algorithms}, vol.~10, no. 114,
  2017.

\bibitem{gb2004}
T.~G. Dietterich, A.~Ashenfelter, and Y.~Bulatov, ``Training conditional random
  fields via gradient tree boosting,'' in \emph{Proceedings of the 21 st
  International Conference on Machine Learning}, Banff, Canada, 2004., 2004.

\bibitem{rnn2010language}
T.~Mikolov, M.~Karafiat, L.~Burget, J.~H. Cernocky, and S.~Khudanpur,
  ``Recurrent neural network based language model,'' in \emph{INTERSPEECH
  2010}, Makuhari, Chiba, Japan, 26-30 September 2010, pp. 1045--1048.

\bibitem{rnnspeech2011}
------, ``Extensions of recurrent neural network language model,'' in
  \emph{2011 IEEE International Conference on Acoustics, Speech and Signal
  Processing (ICASSP)}, 22-27 May 2011.

\bibitem{rnnlocation2016}
Q.~Liu, S.~Wu, LiangWang, and T.~Tan, ``Predicting the next location: A
  recurrent model with spatial and temporal contexts,'' in \emph{Proceedings of
  the 13th AAAI Conference on Artificial Intelligence}, 2016, pp. 194--200.

\bibitem{rnnlocation2017}
D.~Yao, C.~Zhang, J.~Huang, and J.~Bi, ``Serm: A recurrent model for next
  location prediction in semantic trajectories,'' in \emph{Proceedings of the
  2017 ACM on Conference on Information and Knowledge Management}.\hskip 1em
  plus 0.5em minus 0.4em\relax ACM, 2017, p. 2411–2414.

\bibitem{openchalleges2014KDD}
G.~Krempl, I.~Zliobaite, D.~Brzezinski, E.~Hüllermeier, M.~Last, V.~Lemaire,
  T.~Noack, A.~Shaker, S.~Sievi, M.~Spiliopoulou, and J.~Stefanowski, ``Open
  challenges for data stream mining research,'' \emph{ACM SIGKDD Explorations
  Newsletter}, vol.~16, no.~1, pp. 1--10, 2014.

\bibitem{KNNDelayLabel2008}
L.~I. Kuncheva and J.~S. Sanchez, ``Nearest neighbour classifiers for streaming
  data with delayed labelling,'' in \emph{8th IEEE International Conference on
  Data Mining}, 2008, p. 869–874.

\bibitem{TKDEcluster2011}
M.~Masud, J.~Gao, L.~Khan, J.~Han, and B.~M. Thuraisingham, ``Classification
  and novel class detection in concept-drifting data streams under time
  constraints,'' \emph{IEEE Transactions on Knowledge and Data Engineering},
  vol.~23, no.~6, pp. 859 -- 874, 2011.

\bibitem{SemiSupervisedClustering2012}
M.~M. Masud, C.~Woolam, J.~Gao, L.~Khan, J.~Han, K.~W. Hamlen, and N.~C. Oza,
  ``Facing the reality of data stream classification: coping with scarcity of
  labeled data,'' \emph{Knowledge and Information Systems}, vol.~33, no.~1, pp.
  213--244, 2012.

\end{thebibliography}

%




\begin{IEEEbiography}[{\includegraphics[width=1in,height=1.25in,clip,keepaspectratio]{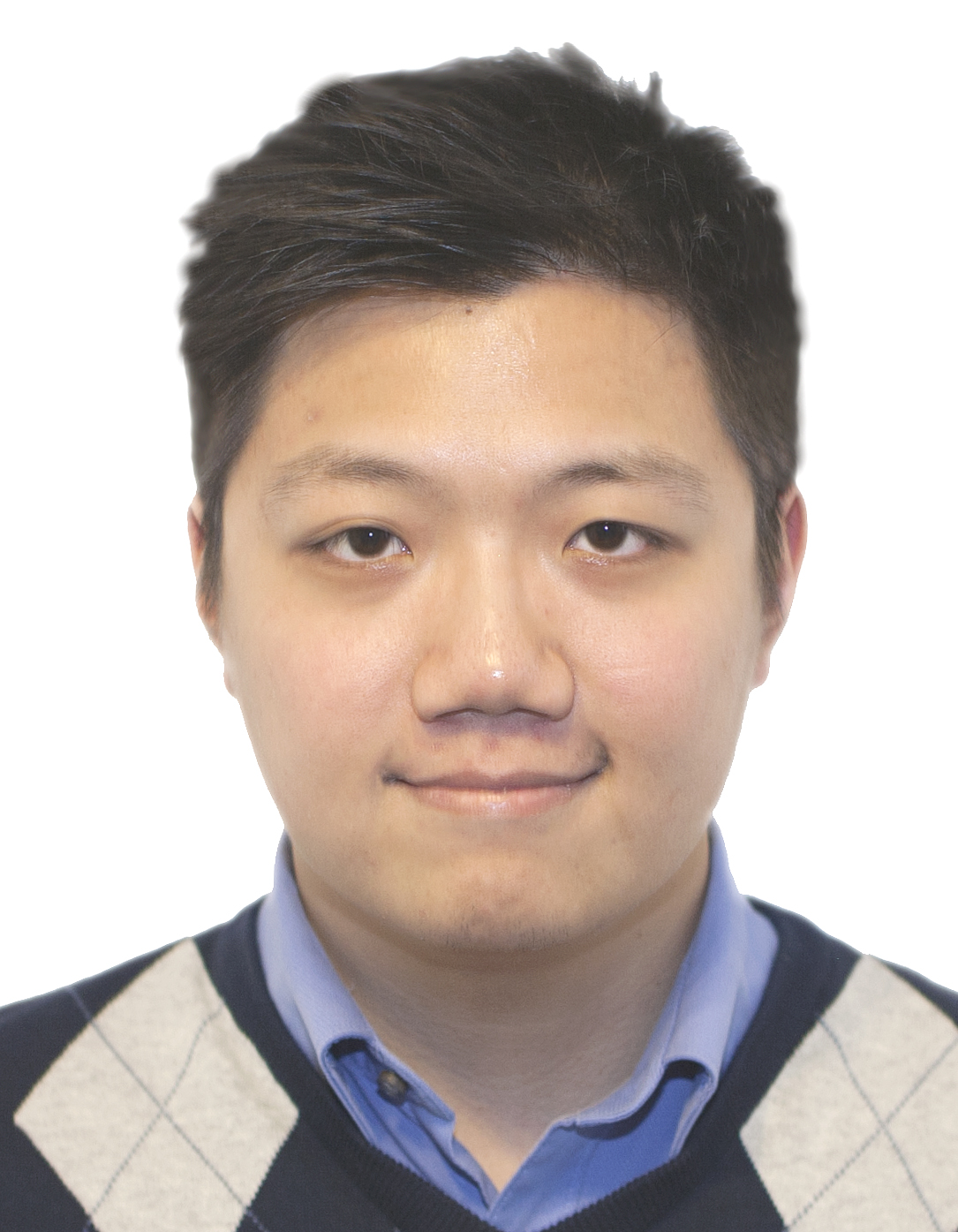}}]{Junxuan Li}
is a Ph.D. candidate in Operations Research from H. Milton Stewart School of Industrial \& Systems Engineering, Georgia Institute of Technology. He was a Microsoft Data Scientist Intern in Summer 2018. His research interests include theories in stochastic programming, dynamic programming, artificial intelligence and machine learning, as well as their applications in engineering, business and healthcare.
\end{IEEEbiography}

\begin{IEEEbiography}[{\includegraphics[width=1in,height=1.25in,clip,keepaspectratio]{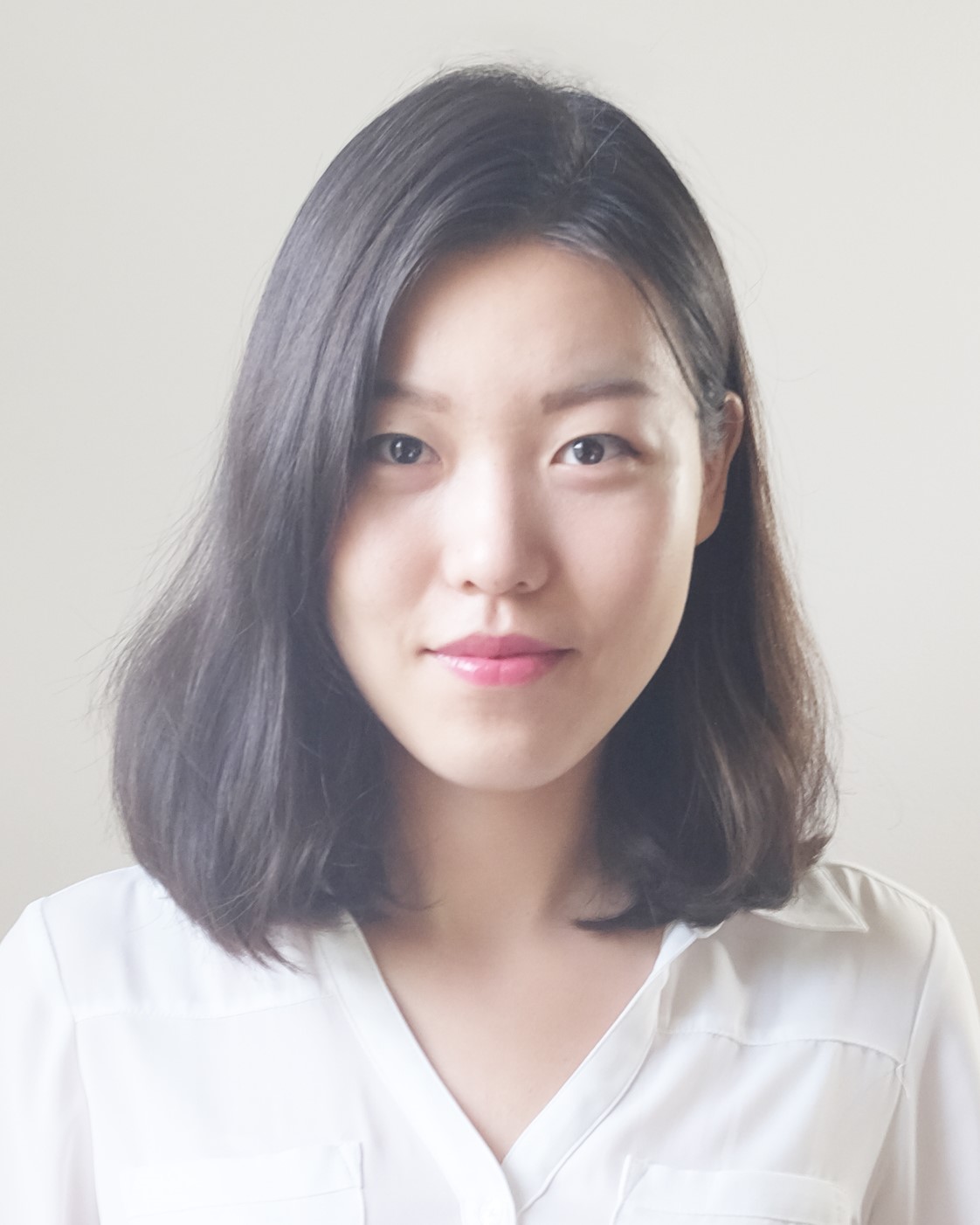}}]{Yifei Ren}
is a Ph.D. candidate in Computer Science from Department of Mathematics and Computer Science, Emory University. She is a research assistant in AIMS research group, with expertise in machine learning, artificial intelligence and deep learning. Her main research interests include pattern recognition and trajectory prediction with spatial-temporal data.
\end{IEEEbiography}

\begin{IEEEbiography}[{\includegraphics[width=1in,height=1.25in,clip,keepaspectratio]{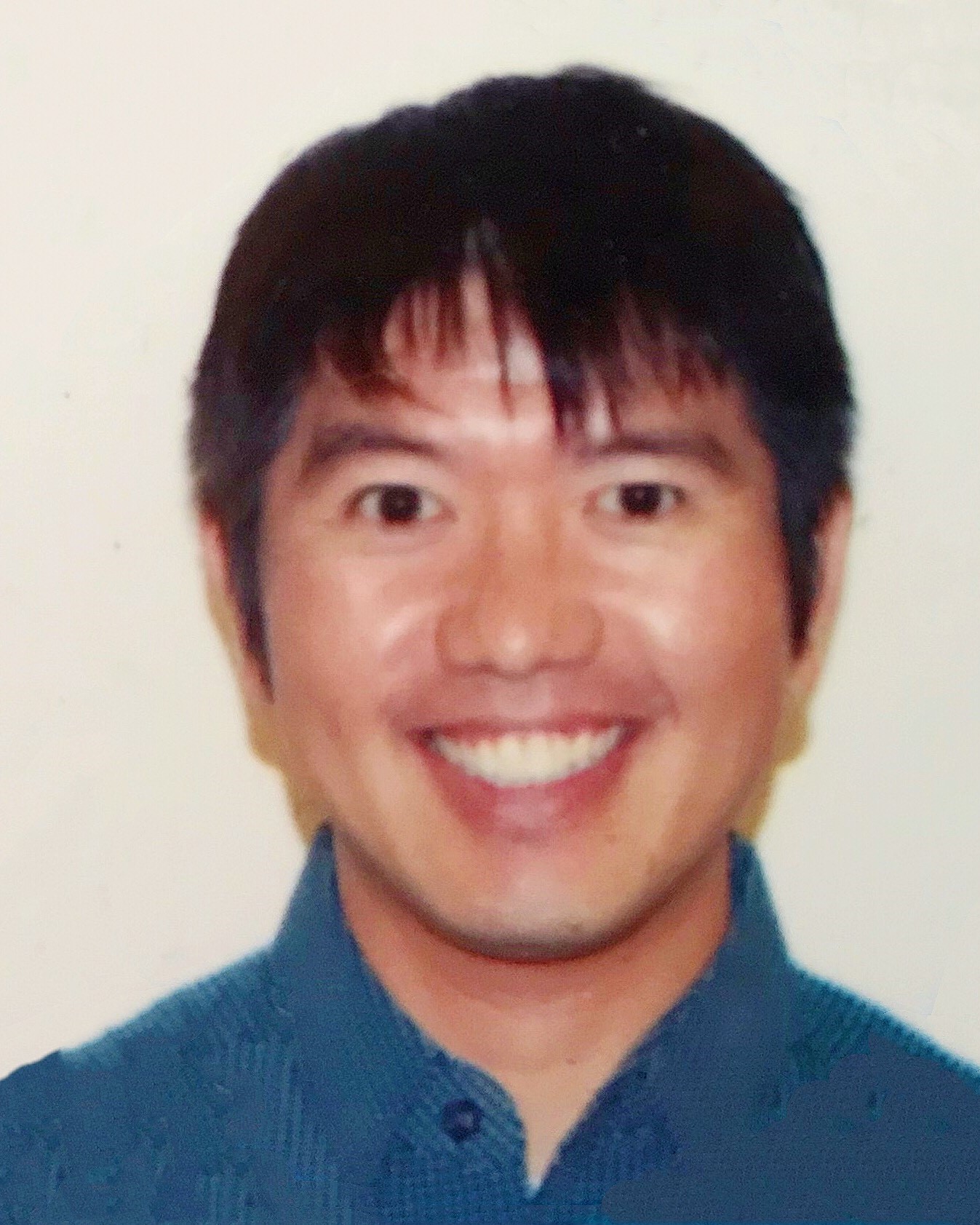}}]{Yung-wen Liu}
Ph.D., is a Senior Data Scientist in Risk Sata Science Team at Microsoft Corporation, expert in data analysis, machine learning modeling, simulation modeling and design of experiment. He received the Ph.D. degree in Industrial Engineering from University of Washington in 2006. Before joining Microsoft in 2017, he was an Associate Professor in the Department of Industrial and Manufacturing Systems Engineering at the University of Michigan - Dearborn since 2006.
\end{IEEEbiography}


\begin{IEEEbiography}[{\includegraphics[width=1in,height=1.25in,clip,keepaspectratio]{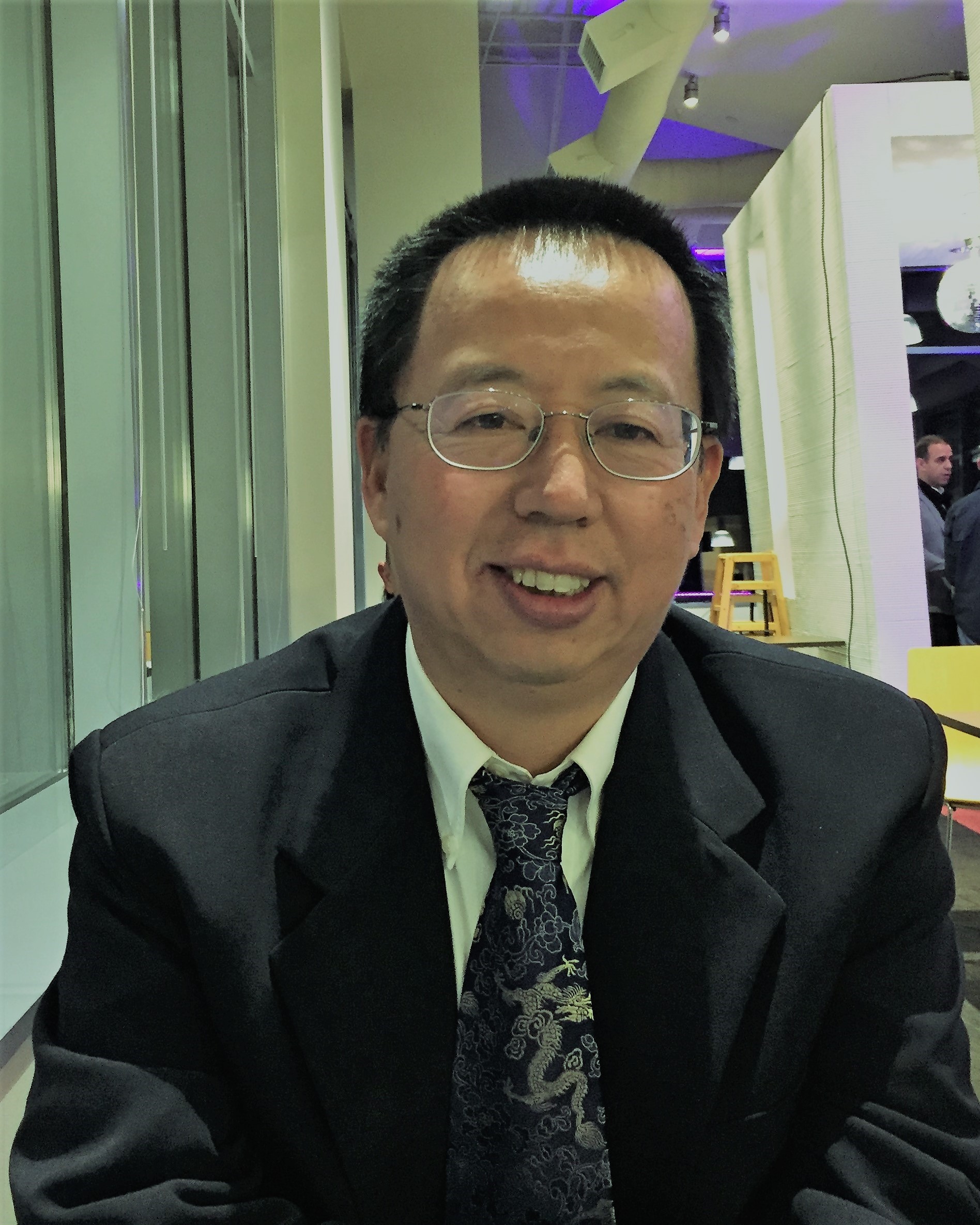}}]{Yuting Jia}
Ph.D., is Principal Data Scientist Lead of risk data science group at Microsoft Corporation. He received the Ph.D. degree in Mathematics from Queen's University in 2001. He leads the Risk Data Science Team to protect against frauds on all purchase transactions happening in Microsoft Universal Store, as well as fighting abuse of web services such as Azure, Bing, and Office. Before working on anti-fraud, he was a Professor in the Department of Mathematics of Hebei Normal University in China.
\end{IEEEbiography}

\begin{IEEEbiography}[{\includegraphics[width=1in,height=1.25in,clip,keepaspectratio]{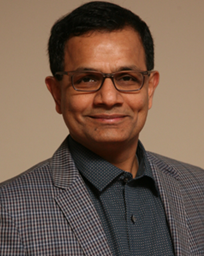}}]{Jay Nanduri}
is a Distinguished Engineer at Microsoft. He received the M.B.A from the Wharton School of the University of Pennsylvania in 2010. Jay has 20+ years’ experience in object oriented distributed systems, secure internet platforms, personalization systems, Ecommerce platforms, ERP products, big data machine learning platforms. His work covers scenarios like modern large-scale system that targets campaigns across many surfaces and Microsoft Loyalty program to modern anti-fraud infrastructure for any purchase to usage across all of Microsoft.
\end{IEEEbiography}




\end{document}